\documentclass[letterpaper, 10 pt, conference]{ieeeconf}  
\usepackage{url}
\usepackage{graphicx}

\usepackage{amsmath,amsfonts}
\usepackage{algorithm}
\usepackage{algpseudocode}

\usepackage{array}
\usepackage[caption=false,font=normalsize,labelfont=sf,textfont=sf]{subfig}
\usepackage{textcomp}
\usepackage{stfloats}
\usepackage{hyperref}
\usepackage{verbatim}
\usepackage{cite}
\usepackage{amssymb}
\usepackage{color}
\usepackage{cleveref}

\usepackage{pifont}
\usepackage{makecell}

\IEEEoverridecommandlockouts                              

\title{\LARGE \bf
MARS-FTCP: Robust Fault-Tolerant Control and Agile Trajectory Planning for Modular Aerial Robot Systems
}

\author{Rui Huang, Zhenyu Zhang, Siyu Tang, Zhiqian Cai,  Lin Zhao
\thanks{This work was supported by the Singapore Ministry of Education Tier 2 AcRF T2EP20123-0037 (Corresponding author: Lin Zhao). Rui Huang,  Zhenyu Zhang, Siyu Tang, and Lin Zhao are with the Department of Electrical and Computer Engineering, National University of Singapore, Singapore 117583, Singapore (email: 
        {ruihuang@u.nus.edu, zhenyuzhang@u.nus.edu, e1352616@u.nus.edu, elezhli@nus.edu.sg}).
        Zhiqian Cai is with the Engineering Design and Innovation Centre, National University of Singapore, Singapore 117583, Singapore (email: 
        {zhiqian@u.nus.edu})
}
}
\begin{document}

\maketitle
\thispagestyle{empty}
\pagestyle{empty}

\begin{abstract}

Modular Aerial Robot Systems (MARS) consist of multiple drone units that can self-reconfigure to adapt to various mission requirements and fault conditions. However, existing fault-tolerant control methods exhibit significant oscillations during docking and separation, impacting system stability. To address this issue, we propose a novel fault-tolerant control reallocation method that adapts to an arbitrary number of modular robots and their assembly formations. The algorithm redistributes the expected collective force and torque required for MARS to individual units according to their moment arm relative to the center of MARS mass. Furthermore, we propose an agile trajectory planning method for MARS of arbitrary configurations, which is collision-avoiding and dynamically feasible. Our work represents the first comprehensive approach to enable fault-tolerant and collision avoidance flight for MARS. We validate our method through extensive simulations, demonstrating improved fault tolerance, enhanced trajectory tracking accuracy, and greater robustness in cluttered environments. The videos and source code of this work are available at \url{https://github.com/RuiHuangNUS/MARS-FTCP/}

\end{abstract}


\section{Introduction}
\begin{figure}[!t]
\centering
\includegraphics[width=3.1 in]{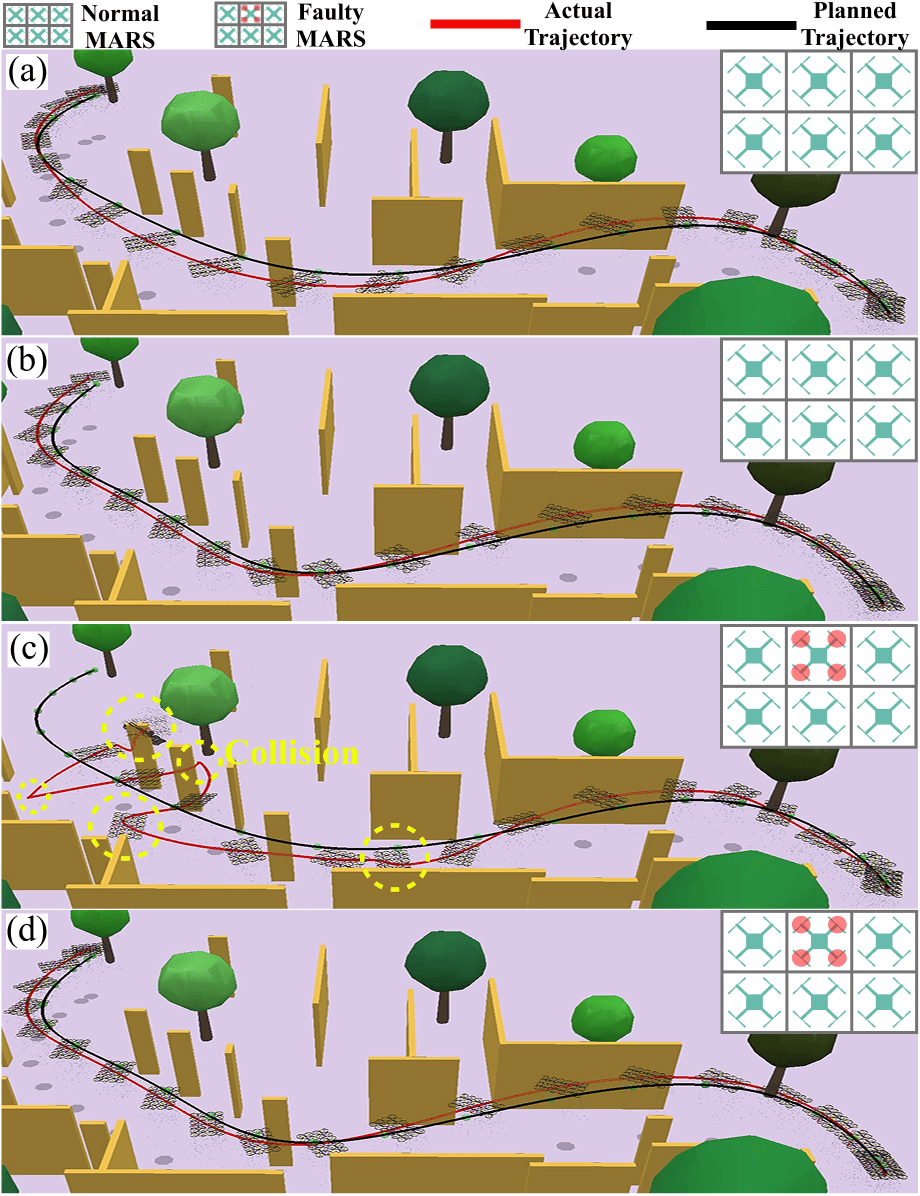}
\vspace{-3mm}
\caption{MARS is tasked with tracking a collision-free trajectory with one faulty unit. The faulty propellers are marked in red. (a) MARS cannot accurately follow the planned trajectory using an existing collision-free trajectory generation method~\cite{wang2024implicit} under a simple PID control and control allocation approach used in previous work~\cite{huang2024adaptive}. (b) MARS can track the planned trajectory more accurately using our dynamics-aware planning method. (c) MARS fails to track the trajectory planned with~\cite{wang2024implicit} under our proposed fault-tolerant control. (d) MARS can track the trajectory planned with our proposed method relatively accurately under our proposed fault-tolerant control.}
\label{fig:MARS}
\vspace{-7mm}
\end{figure}

Modular Aerial Robot Systems (MARS) \cite{oung2014distributed, su2024flight} consist of flying units capable of freely maneuvering in three-dimensional space. They can reconfigure the assembly formation in-flight to facilitate navigating through cluttered and confined environments~\cite{Yi-RSS-23,xu2023finding,jia2023aerial}. Despite the promising applications, agile trajectory planning and control of MARS is rarely explored in the literature that taps into such flexibility. Simple applications of existing generic planning and control algorithms typically lead to poor tracking performances. For example, MARS under a simple PID controller cannot accurately track the collision-free trajectory generated by previous work~\cite{wang2024implicit} (See Fig.~\ref{fig:MARS}(a)).

Besides, aerial robots are vulnerable to various faults, including multiple propeller failures that jeopardize the control performance and flight safety \cite{wang2017adaptive, saied2023review}. This can severely degrade the control performance of MARS. Achieving fault tolerance in various robotic swarms is a challenging problem that has garnered significant attention in recent research \cite{saldana2018modquad,gandhi2020self,huang2024adaptive}. Existing works on MARS primarily focus on docking and separation \cite{li2019modquad,zhang2024design,sugihara2024beatle}, coordinated flight\cite{huang2024adaptive}, and self-reconfiguration\cite{gandhi2020self,Huang2025Robust}. However, the shape of MARS may not be well-suited for navigating through narrow urban environments, which limits its adaptability in load-carrying and shape-transforming transportation tasks. To enhance its versatility and operational efficiency, a method capable of planning arbitrary MARS configurations is required.

Existing studies \cite{zhang2023continuous,wang2024implicit} on collision-free trajectory generation primarily focus on computing two-dimensional and three-dimensional trajectories for arbitrary shapes in complex environments. However, these methods do not consider asymmetrically distributed control outputs, such as thrusters and motors commonly used in aerial and underwater robotics. Consequently, the control performance varies across different directions, causing deviations between the planned collision-free trajectory and the system's actual tracking capability. Fig.~\ref{fig:MARS}(c) shows that MARS fails
to track the trajectory planned with \cite{wang2024implicit} under a fault-tolerant controller (FTC) proposed in this paper, as the former does not account for the dynamics in their planning.

To enhance the in-flight safety of MARS, we propose a novel dynamics-aware collision-free trajectory planning algorithm that incorporates FTC during obstacle avoidance. As a preview of our results, Fig.~\ref{fig:MARS}(b) demonstrates that MARS can track the planned trajectory more accurately using our proposed dynamics-aware planning method, and Fig.~\ref{fig:MARS}(d) shows the performance under our proposed FTC scheme. Compared to Fig.~\ref{fig:MARS}(a) and (c), these results highlight the importance of incorporating the dynamics into the trajectory planning and the control design to achieve robust and agile control of MARS.

Moreover, recent work \cite{gandhi2020self} proposed a self-reconfiguration technique for inter-module failures, designed to enable mission continuation while mitigating the effects of rotor failures. This method reduces the impact of failures on the trajectory by employing a mixed-integer linear program to determine optimal module placement. However, it primarily focuses on connecting a functional module to a unit experiencing a partial failure, such as the loss of a single rotor in a quadcopter, and does not address scenarios involving complete module failures or multiple drone unit failures. Furthermore, when a module docks or separates, the remaining MARS experiences severe oscillations, significantly compromising flight safety. Additionally, the study does not provide details on the fault-tolerant control method employed. 

In summary, our key contributions include:
\begin{enumerate}
    \item We propose a novel fault-tolerant control reallocation method that adapts to arbitrary number of modular robots and their assembly formations. Previous method \cite{gandhi2020self}  exhibit significant oscillations before and after docking and separation. We successfully eliminate such dangerous transitions by the proposed control reallocation. 
    \item We propose an agile trajectory planning method for MARS of arbitrary configurations, which is collision-avoiding and dynamically feasible. In particular, we enhance the A* algorithm to compute dynamically feasible nodes by maximizing the control efficiency of the longitudinal flight direction of MARS, which significantly improves its tracking performance. Our work represents the first comprehensive approach to enable fault-tolerant and collision avoidance flight for MARS.
    \item Lastly, our approach achieves superior tracking performance in the presence of faults, more accurate and stable than the baseline algorithm~\cite{wang2024implicit}.
\end{enumerate}


\section{Fault tolerance adaptive configuration control of MARS}
In this section, we propose a fault-tolerant reallocation method for MARS that redistributes total force and torque after a failure, ensuring a symmetric force distribution and significantly enhancing fault tolerance. We assume that the structural configuration remains unchanged before and after the failure, which means that no faulty unit can be discarded from the MARS. 

\subsection{Units failure}

When a failure or degradation occurs, the affected unit experiences a reduction in total thrust. In the case of a complete failure, the faulty unit can be approximated as an additional load. To compensate for the unbalanced thrust caused by faulty units, a redistribution of total thrust is required. We constrain the torque generated by each unit to prevent undesired fault-induced torque while assuming that the total force and torque satisfy the assembly force requirements. The optimization problem to determine the thrust coefficient parameters is defined as follows:
\begin{subequations} 
\begin{align}
& \min_{\boldsymbol{u}_a} Var(\boldsymbol{u}_a) \\
{\rm s.t.}\ 
& \Sigma u_{a,i}\cdot \mathbf{P}_i=0\\
& \Sigma u_{a,i}=F_{assembly}\\
&  \boldsymbol{\tau}^{\pm}=\mathbf{M}^{\pm}_{assembly}
\end{align}
\end{subequations}
where $\boldsymbol{u}_a=[F_1,\cdots,F_i]$ represents the control output of the MARS, and $F_i$ denotes the force produced by the $i$-th normal unit. The function $\text{Var}(\cdot)$ represents the variance. The unit position in the assembly coordinate frame is given by $\mathbf{P}_i=[x_i,y_i]$. The desired force and net torque of the MARS are denoted as $F_{assembly}$ and $\mathbf{M}^{\pm}_{assembly}=[M_x^+,M_x^-,M_y^+,M_y^-]$, respectively. The current torque generated by the thrusters is represented as $\boldsymbol{\tau}^{\pm}=[\tau_x^+,\tau_x^-,\tau_y^+,\tau_y^-]$, which is computed as:
\begin{align}
\boldsymbol{\tau}^\pm =
\begin{bmatrix}
\min(  \begin{bmatrix}
        \boldsymbol{x}_{a} & 
       \boldsymbol{y}_{a} 
    \end{bmatrix}^T,\mathbf{0}_{2i\times1})
    \boldsymbol{u}_{a} \\
\max(\begin{bmatrix}
        \boldsymbol{x}_{a} &
       \boldsymbol{y}_{a} 
    \end{bmatrix}^T ,\mathbf{0}_{2i\times1}) 
    \boldsymbol{u}_{a} 
\end{bmatrix}
\label{eq:tau}
\end{align}
where, min($\cdot$) and max($\cdot$) operate element-wise, selecting the minimum and maximum values for each corresponding element of the input vectors.
$\boldsymbol{x}_a = \begin{bmatrix}x_{1}, & \cdots, & x_{i}\end{bmatrix}^T$ and $\boldsymbol{y}_a= \begin{bmatrix}y_{1}, & \cdots, & y_{i}\end{bmatrix}^T$ denote the unit positions in the MARS coordinate frame. 

\subsection{Rotors failure}

Rotor failure is characterized by a condition in which the affected unit continues to generate partial thrust and torque but fails to achieve the performance of a fully functional unit. In such cases, the entire system must balance the collective thrust and net torque. To effectively utilize the remaining lift and torque produced by faulty units, two potential approaches can be employed: An intuitive method to reallocate force and torque of the assembly is to modify the coefficients of the remaining normal quadrotors, while keeping all control inputs of the faulty unit unchanged. The new collective thrust and torque provided by all normal units can be calculated as follows: partial reallocation and full reallocation.

\subsubsection{Partial Reallocation}

An intuitive method to reallocate force and torque of the assembly is to modify the coefficients of the remaining normal quadrotors, while keeping all control inputs of the faulty unit unchanged. The new collective thrust $F_{normal}^{\chi} $ and torque $\mathbf{M}_{normal}^{\chi}$ provided by all normal units can be calculated as follows:
\begin{equation}
    F_{normal}^{\chi} = {\sum}_{i=1}^{k}   F_i  + \Delta F_{failed} \label{eq:F_assembly}
\end{equation}
\begin{equation}
   \mathbf{M}_{normal}^{\chi} = {\sum}_{i=1}^{k}  \mathbf{M}_i + {\Delta} \mathbf{M}_{failed}
     \label{eq:M_assembly}
\end{equation}
where $k$ is the number of normal units. $\Delta F_{failed}$ and $\Delta\mathbf{M}_{failed}$ denote the loss of thrust and torque due to rotor failure, respectively, and can be calculated using the control efficiency matrix. Then, the angular velocity $\boldsymbol{\omega}_{i}$ can be determined according to the adaptive configuration control method without failure \cite{huang2024adaptive}.
\begin{equation}
    \begin{bmatrix}
        \omega^2_{i,1} & \cdots 
      &  \omega^2_{i,n_r}
    \end{bmatrix}^T
    =
    \mathbf{P}_{n_r} \mathbf{E}_i
    \begin{bmatrix}
        F_{normal}^{\chi} \\
        \mathbf{M}_{normal}^{\chi}
    \end{bmatrix}
     \label{eq:allocation_calculate}
\end{equation}
where $\mathbf{P}_{n_r}$ denotes the original control allocation matrix of a multirotor with $n_r$ propellers, and $\mathbf{E}_i = \text{diag} \left([n^{-1}, | \mathbf{P}_i+\mu| \cdot J_{assembly}  / |\sum_{i}^{n} (\mathbf{P}_i+\mu) \cdot  J_{unit}| ] \right) \in \mathbb{R}^{4 \times 4}$ defines the adaptive configuration control allocation parameters. Here, $J_{assembly},J_{unit}$ represent the moment of inertia for MARS and individual units, respectively. $\mu$ is a small constant introduced to prevent division by zero.

\subsubsection{Full Reallocation}

The performance of partial reallocation may deteriorate when the total number of units in the assembly is limited. To address this limitation, an alternative strategy is proposed, employing a two-stage approach to redistribute the force and torque of the MARS when only a few units remain operational. In the first stage, the faulty unit is stabilized by maintaining balance in the pitch, roll, and yaw directions. Subsequently, the remaining units compensate for the loss of force and torque. To determine the internal force and torque distribution, the following optimization problem is formulated: 
\begin{subequations} 
\begin{align}
& \min_{f_{i,j}} Var(f_{i,j}) \\
{\rm s.t.}\ 
& \Sigma f_{i,j}\cdot \mathbf{P}_{i,j}=0\\
& \Sigma \boldsymbol{\tau}_{i,j}^{\pm}=\boldsymbol{\tau}_{i}^{\pm}
\end{align}
\end{subequations}
where $f_{i,j}$ represents the thrust of the faulty unit that needs to be compensated by other parts of the assembly. The propeller positions in the assembly coordinate frame are given by $\mathbf{P}_{i,j} = [x_{i,j}, y_{i,j}]$. The terms ${\tau}_{i,j}^{\pm}$ and ${\tau}_{i}^{\pm}$ describe the net torque of the propeller and the unit, respectively. The loss of force and torque for the faulty unit can then be determined, and the assembly's angular velocity calculations follow a similar approach to \eqref{eq:F_assembly}, \eqref{eq:M_assembly}, and \eqref{eq:allocation_calculate}.

\section{Collision-free Trajectory Planning}
\begin{figure}[!t]
\centering
\includegraphics[width=3.4in]{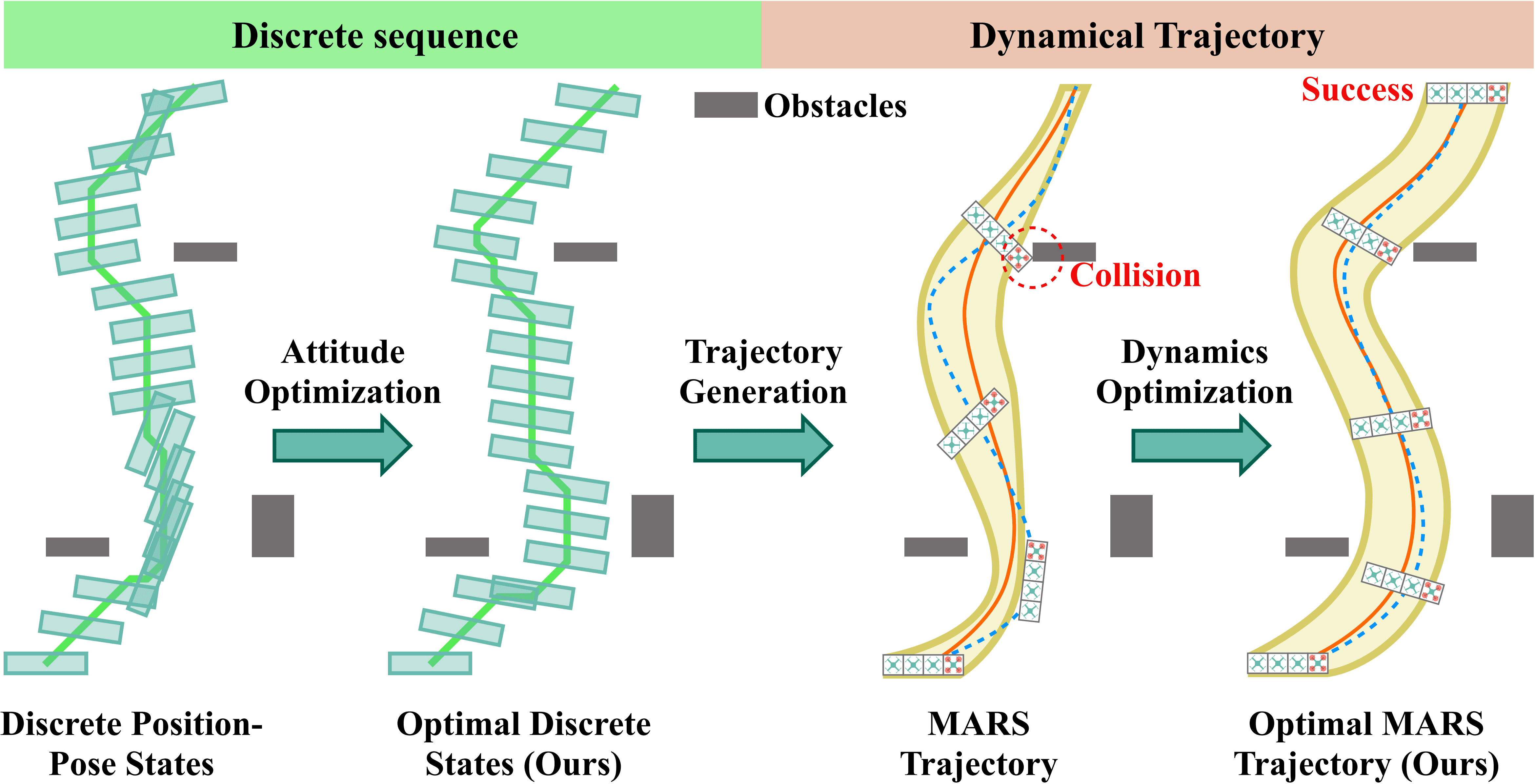}
\vspace{-6mm}
\caption{Illustration of  dynamics-aware collision-free trajectory generation process. Initially, a discrete sequence is computed using the asymmetric A* algorithm. This sequence is then refined through attitude optimization. Subsequently, a trajectory is generated based on the optimized discrete states. However, the initial trajectory may lead to collisions due to dynamic constraints. To address this, a dynamics optimization step is applied, resulting in an optimal MARS trajectory that successfully avoids obstacles while enhancing the control performance of MARS. }
\label{fig:trajectory_generation}
\vspace{-7mm}
\end{figure}
In this section, we introduce a dynamics-aware collision-free trajectory planning algorithm. Unlike previous work \cite{wang2024implicit}, which focuses solely on motion continuity constraints while overlooking the fact that the control performance of arbitrarily shaped robotic systems varies across different directions in real-world scenarios, our approach explicitly accounts for these anisotropies. The trajectory generation workflow is illustrated in Fig.~\ref{fig:trajectory_generation}. First, a discrete sequence is generated using the asymmetric A* algorithm \cite{wang2024implicit}. To enhance trajectory tracking feasibility, we apply attitude optimization to the discrete position-pose states, enabling robotic systems to fully utilize their control capabilities. Next, the discrete sequence is converted into an initial trajectory. Finally, a continuous dynamical trajectory for collision-free trajectory will be generated after dynamics optimization. 

\begin{algorithm}
\label{algorithm:A*}
\small
\caption{Dynamically Feasible A* Algorithm}
\begin{algorithmic}[1]
\State \textbf{Input:} $N_{start}, N_{end}, \Phi$
\State \textbf{Initialize:} Open set $\mathcal{O} \leftarrow \{N_{start}\}, $Closed set $\mathcal{C} \leftarrow \{\}$
\State \textbf{Set} $\mathcal{G}(N_{start}) = 0$, $\mathcal{F}(N_{start}) = \mathcal{H}(N_{start},N_{end})$
\While{ $\mathcal{O} \neq \emptyset$ }
    \State $N_{curr} \leftarrow$ node in $\mathcal{O}$ with lowest $\mathcal{F}$
    \If{$N_{curr} == N_{end}$}
        \State \textbf{Return} discrete sequence
    \EndIf
    \State $\mathcal{O} \leftarrow \mathcal{O} \setminus \{N_{curr}\}, \mathcal{C} \leftarrow \mathcal{C} \cup \{N_{curr}\}$
    \For{each neighbor $N_{near}$ of $N_{curr}$}
        \If{$N_{near} \notin C$ \textbf{and} \Call{CollisionCheck}{$N_{near}$}}
            \State $\mathcal{G}_{temp} = \mathcal{G}(N_{curr}) + distance(N_{curr}, N_{near})$
            \State \Comment{Compute path cost}
            \If{$N_{near} \notin \mathcal{O}$ or $\mathcal{G}_{temp} < \mathcal{G}(N_{near})$}
                \State $\mathcal{G}(N_{near})=\mathcal{G}_{temp}$ 
                \State $\Phi^{*} = \Call{OptimalAttitude}{N_{near}}$ \eqref{eq:phi}
                \State $\mathcal{F}(N_{near}) = \mathcal{G}(N_{near}) + \mathcal{H}(N_{near},N_{end}) + \mathcal{J}(\Phi^*, \Phi_{N_{near}})$ \eqref{eq:node_cost}
                \State $N_{near}.parent\leftarrow N_{curr}  $
                \State  $\mathcal{O} \leftarrow \mathcal{O} \cup \{N_{near}\} $
            \EndIf
        \EndIf
    \EndFor
\EndWhile
\State \textbf{Return} failure
\end{algorithmic}
\end{algorithm}

\subsubsection{Discrete sequence generation}
Previous work \cite{wang2024implicit} generated a discrete sequence using the A* algorithm, which asymmetrically expands neighboring nodes in both position and attitude dimensions. However, this method does not consider system dynamics, which may render the generated path infeasible. As a result, the method produces an idealized continuous trajectory that cannot be accurately followed by the robotic system, leading to infeasible motion.

Therefore, we modified the cost function during the expansion of neighboring nodes, formulating the \textbf{dynamically feasible A* algorithm}, which generates a discrete sequence optimized for dynamics feasibility. Our algorithm begins the search from the start node $N_{start}$, iteratively selecting the current node $N_{curr}$ with the lowest total cost $\mathcal{F}(\cdot)$ within the open set $\mathcal{O}$. For all neighboring nodes $N_{near}$ of the current node, the algorithm identifies the collision-free neighbor with the lowest path cost, adds it to the open set, and updates its total cost, which includes the heuristic cost $\mathcal{H}(\cdot)$ and the attitude cost $\mathcal{J}(\cdot)$.

\begin{figure*}[!t]
\centering
\includegraphics[width=6.8in]{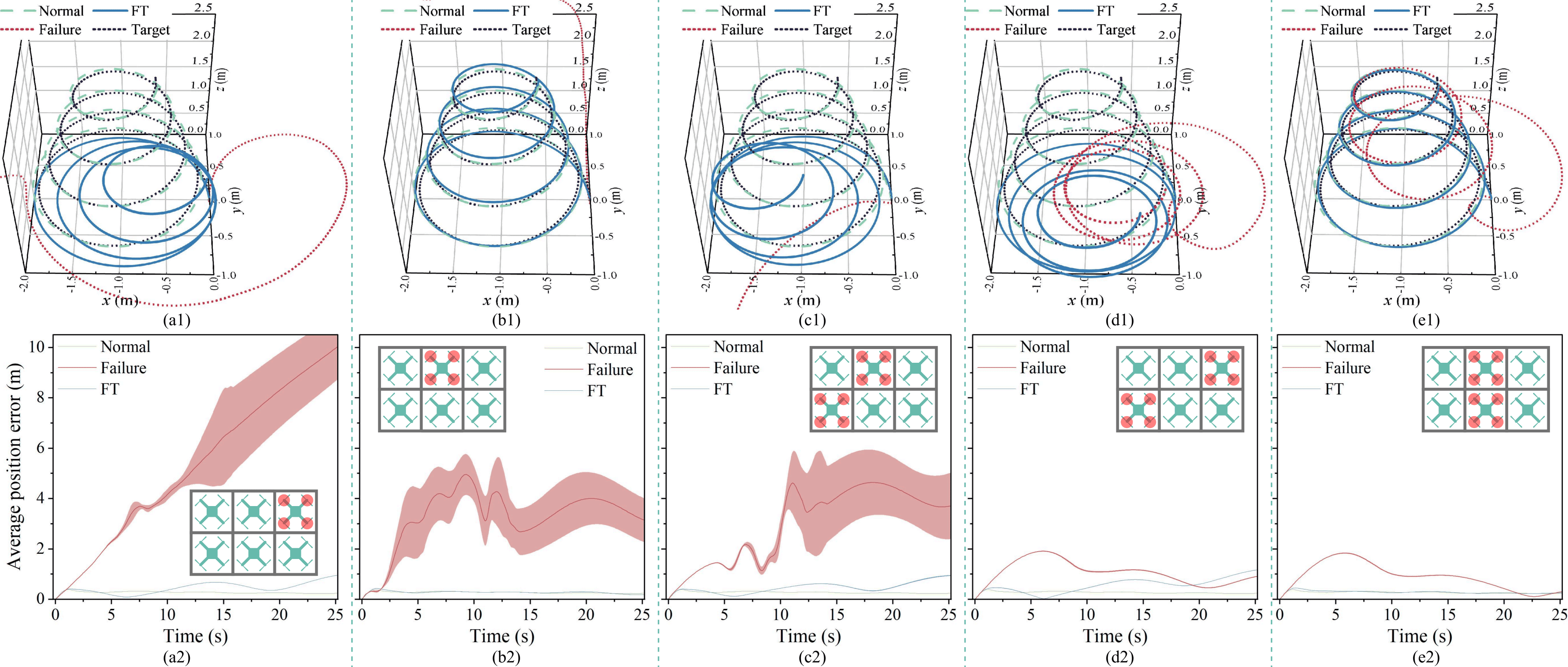}
\vspace{-3mm}
\caption{(a1)–(e1): Trajectory tracking results with and without FTC. (a2)–(e2): Average position errors across ten repeated trials under different fault scenarios.  
(a), (b): FTC applied to a $3 \times 2$ assembly with one faulty unit. (c), (d), (e): FTC applied to a $3 \times 2$ assembly with two faulty units. The faulty propellers are marked red.}
\label{fig:Complete Failure of One and Two Units}
\vspace{-7mm}
\end{figure*}

To obtain attitude cost, we formulate the optimization problem for general robotic systems with the following equations:
\begin{equation}
\Phi^{*}=\arg\min_{\Phi}  \boldsymbol{\tau}_\pm ^\text{max}\cdot \mathbf{C}_\tau
     \label{eq:phi}
\end{equation}
where the objective function \(\boldsymbol{\tau}_\pm^\text{max} \cdot \mathbf{C}_\tau\) represents the net control output. Here, we consider torque along the \(x\)- and \(y\)-axes in both positive and negative directions. The weight vector is defined to unify the positive and negative signs as  $\mathbf{C}_\tau = \begin{bmatrix}1 & 1 & -1 & -1\end{bmatrix}^T$. The vector $\Phi = [\phi, \theta, \psi]$ represents the rotation angles. The term \(\boldsymbol{\tau}_\pm^\text{max}\) denotes the maximum moment generated by all units at the optimal attitude $\Phi^{*}$, defined as  $\boldsymbol{\tau}_\pm^\text{max} = \begin{bmatrix}
\tau_{y^-}^\text{max}, & \tau_{x^+}^\text{max}, & \tau_{y^+}^\text{max}, & \tau_{x^-}^\text{max}
\end{bmatrix}.$ Here, we use MARS as an example to calculate the torque components:
\begin{align}
\boldsymbol{\tau}_\pm ^\text{max}=
\begin{bmatrix}
\min(R(\phi) R(\theta) R(\psi)\begin{bmatrix}
        \boldsymbol{x}_{a} & 
       \boldsymbol{y}_{a} 
    \end{bmatrix}^T,0)
    \boldsymbol{u}_{a}^\text{max} \\
\max( R(\phi) R(\theta) R(\psi)   \begin{bmatrix}
        \boldsymbol{x}_{a} &
       \boldsymbol{y}_{a} 
    \end{bmatrix}^T,0) 
    \boldsymbol{u}_{a}^\text{max}   
\end{bmatrix}
\end{align}
where $R(\phi)$, $R(\theta)$, and $R(\psi)$ represent the rotation matrices about the $z$-, $y$-, and $x$-axes, respectively. The maximum force contributions for each unit and the entire assembly are defined as:  
$\boldsymbol{u}_{\mathcal{M},i}^\text{max} = \begin{bmatrix}
f_{i1}^\text{max}, & f_{i2}^\text{max}, & f_{i3}^\text{max}, & f_{i4}^\text{max}
\end{bmatrix}^T, \quad
\boldsymbol{u}_{a}^\text{max} = \begin{bmatrix}
\boldsymbol{u}_{\mathcal{M},1}^\text{max}, & \cdots, & \boldsymbol{u}_{\mathcal{M},n}^\text{max}
\end{bmatrix}^T \in \mathbb{R}^{4n}$.

The optimization problem maximizes the control capacity of the robotic system using the maximum torque along the $x$-, $y$- and $z$-axes. To address the feasibility of the following trajectory tracking task, the cost function of the dynamically feasible A* algorithm is defined as:
\begin{subequations}
\begin{align}
& \min_{N_{node}} \text{Cost}_{\mathcal{A}^* } = \mathcal{G}_p + \mathcal{H}_h +  \mathcal{J}_\Phi
\label{eq:node_cost}\\
& \mathcal{J}_\Phi = \mathcal{L}_{\Phi} \left| \Phi^{*} - \Phi^{N_{node}} \right|
\end{align} 
\end{subequations}
where $\mathcal{G}_p$ and  $\mathcal{H}_h$ represent the path cost and the heuristic cost,  $\mathcal{J}_\Phi$ denotes the attitude deviations between node $\Phi^{N_{node}}$ and optimal attitude $\Phi^*$ with the weight coefficient $\mathcal{L}_{\Phi}$.  

The discrete sequence generation produces a series of high-dimensional collision-free and dynamically feasible nodes that contain both position and orientation information. This sequence is defined as follows:
\begin{equation}
\mathcal{S}_{\mathcal{A}^*} = \{N_{{node}}^i : (x^i, y^i, z^i, \Phi^i) \in \text{SE}(3) \}
\end{equation}
\subsubsection{Dynamical trajectory}

\begin{figure*}[!t]
\centering
\includegraphics[width=6.8in]{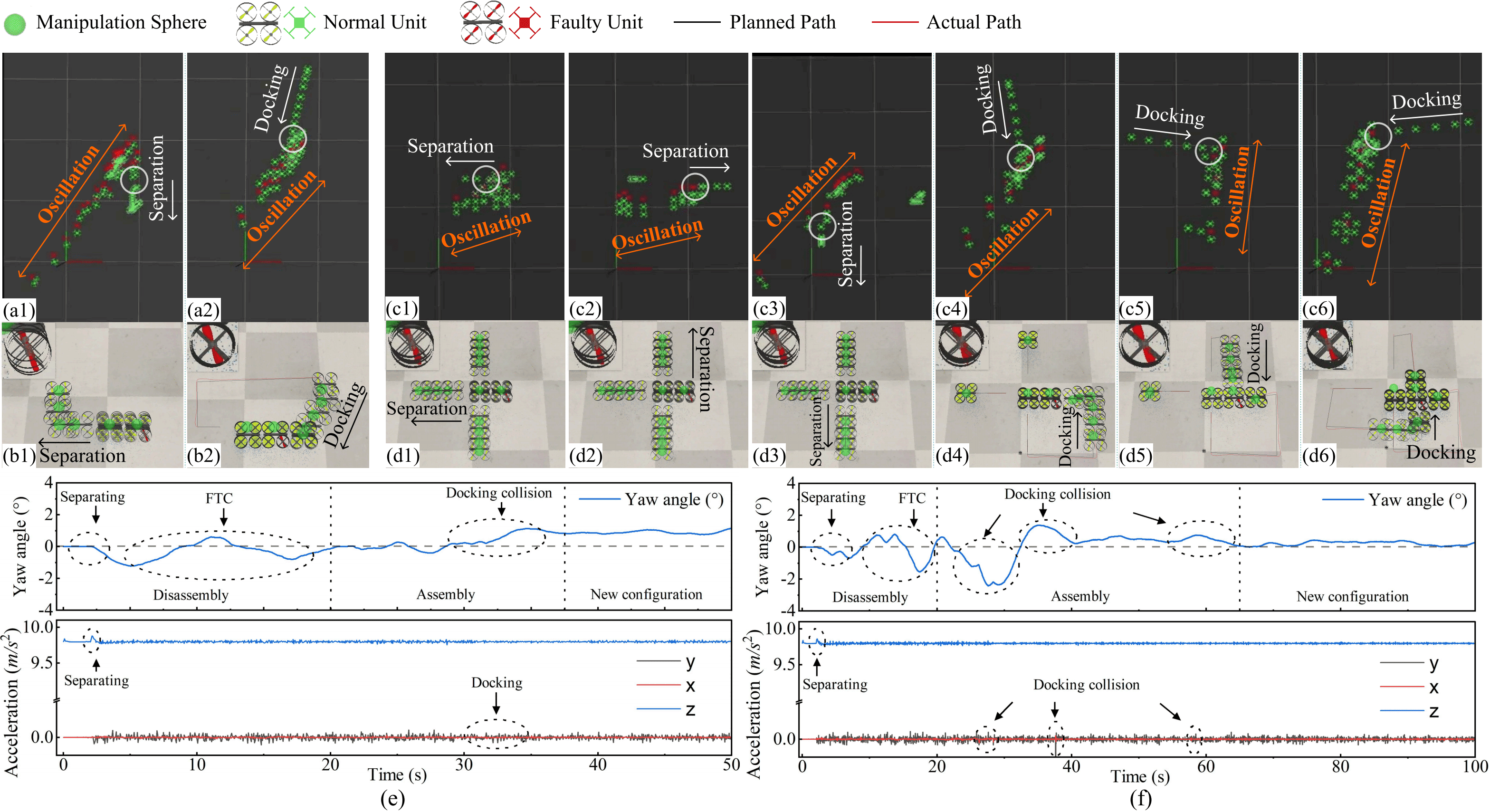}
\vspace{-5mm}
\caption{
FTC performance during self-reconstruction in $3\times1$ and $3\times3$ plus assemblies. Faulty propellers are highlighted in red, while orange arrows indicate the range of oscillation.  
(a) $3\times1$ assembly (comparison method).  
(b) $3\times1$ assembly (Ours).  
(c) $3\times3$ plus assembly (comparison method).  
(d) $3\times3$ plus assembly (Ours).  
(e) and (f) Yaw angle and IMU-measured accelerations of the faulty unit during self-reconfiguration FTC in $3\times1$ and $3\times3$ plus assemblies, respectively. Note that (a) and (c) are sequences from the original video in \cite{gandhi2020self}, which uses the Crazyflie simulator. Moreover, the previous work \cite{gandhi2020self} does not specify the fault-tolerant control method used, and the code is not open-source; therefore, we are unable to evaluate it within the same simulator environment.
}
\label{fig:Oscillation fig}
\vspace{-5 mm}
\end{figure*}
Following the dynamically feasible A* algorithm in the discrete sequence generation, we utilize swept volume signed distance field (SVSDF)  \cite{wang2024implicit}  to generate the continuous trajectory through the discrete sequences $\mathcal{S}_\mathcal{A^*}$. Subsequently, a dynamic trajectory is formulated to ensure that the system can perform the tracking task.

However, previous work \cite{wang2024implicit} considers kinematic factors as the dynamic feasibility term, resulting in merely continuous motion while overlooking the fact that the control performance of robotic systems varies in different directions. To address this limitation, we introduce an optimal attitude term within the dynamic feasibility term to prevent unreasonable attitude variations along the trajectory and enhance overall stability, the cost function in dynamical trajectory is as follows:
\begin{equation}
\min_{\mathbf{c},\mathbf{T}} \text{Cost} =  \lambda_m J_m + \lambda_t J_t + \lambda_o \mathcal{G}_o+\lambda_d \mathcal{G}_d
\end{equation}
where $\mathbf{c}$ and $\mathbf{T}$ denote the waypoints and timestamps, respectively. The terms $J_m$, $J_t$, $\mathcal{G}_o$, and $\mathcal{G}_d$ represent the smoothness, total time, obstacle penalty, and dynamic penalty, respectively. Specifically, $\mathcal{G}_d$ is defined as follows:
\begin{subequations}
\begin{align}
& \mathcal{G}_d(t) = \lambda_v \mathcal{G}_v + \lambda_a \mathcal{G}_a + \lambda_j \mathcal{G}_j + \lambda_{\Phi} \mathcal{G}_\Phi \\
& \mathcal{G}_\Phi = {\sum}_{i=timestamp}^{T} \left| \Phi_{t_{i}}^* - \Phi_{t_{i}} \right|
\end{align}
\end{subequations}
where the term $\mathcal{G}_v$, $\mathcal{G}_a$, and $\mathcal{G}_j$ represent the magnitudes of velocity, acceleration, and jerk, while $\mathcal{G}_\Phi$ denotes the optimal attitude deviation over the entire trajectory. Due to the inclusion of $\mathcal{G}_{\Phi}$, our approach achieves a balance between pure obstacle avoidance and dynamic feasibility.

\section{Evaluation}

We employ a high-fidelity quadrotor model in CoppeliaSim~\cite{coppeliaSim} for simulation and evaluation. Various experiments are conducted to assess fault-tolerant trajectory tracking, rotor failure and degradation, self-reconfiguration, and collision-free trajectory planning.

\subsection{Fault tolerance trajectory tracking}
\subsubsection{Complete Failure of One and Two Units}

To evaluate the performance of the proposed method, a spiral trajectory tracking experiment is conducted. Fig.~\ref{fig:Complete Failure of One and Two Units} presents the results of trajectory tracking along with the average position errors over 10 repeated trials. As shown in Fig.~\ref{fig:Complete Failure of One and Two Units}(a) and (b), when fault-tolerant control is enabled, the assembly remains capable of effectively tracking the spiral direction. Although the tracking trajectory deviates from the target due to increased fault-induced torque when faulty units are positioned far from the center, previous works~\cite{gandhi2020self, Huang2025Robust} have proposed effective approaches to mitigate these drawbacks. In contrast, without fault-tolerant control, the assembly quickly diverges from the intended trajectory after take-off and fails to complete the tracking task. These results demonstrate that the proposed fault-tolerant control method effectively redistributes power to handle fault scenarios. 

\begin{figure*}[!t]
\centering
\includegraphics[width=6.8in]{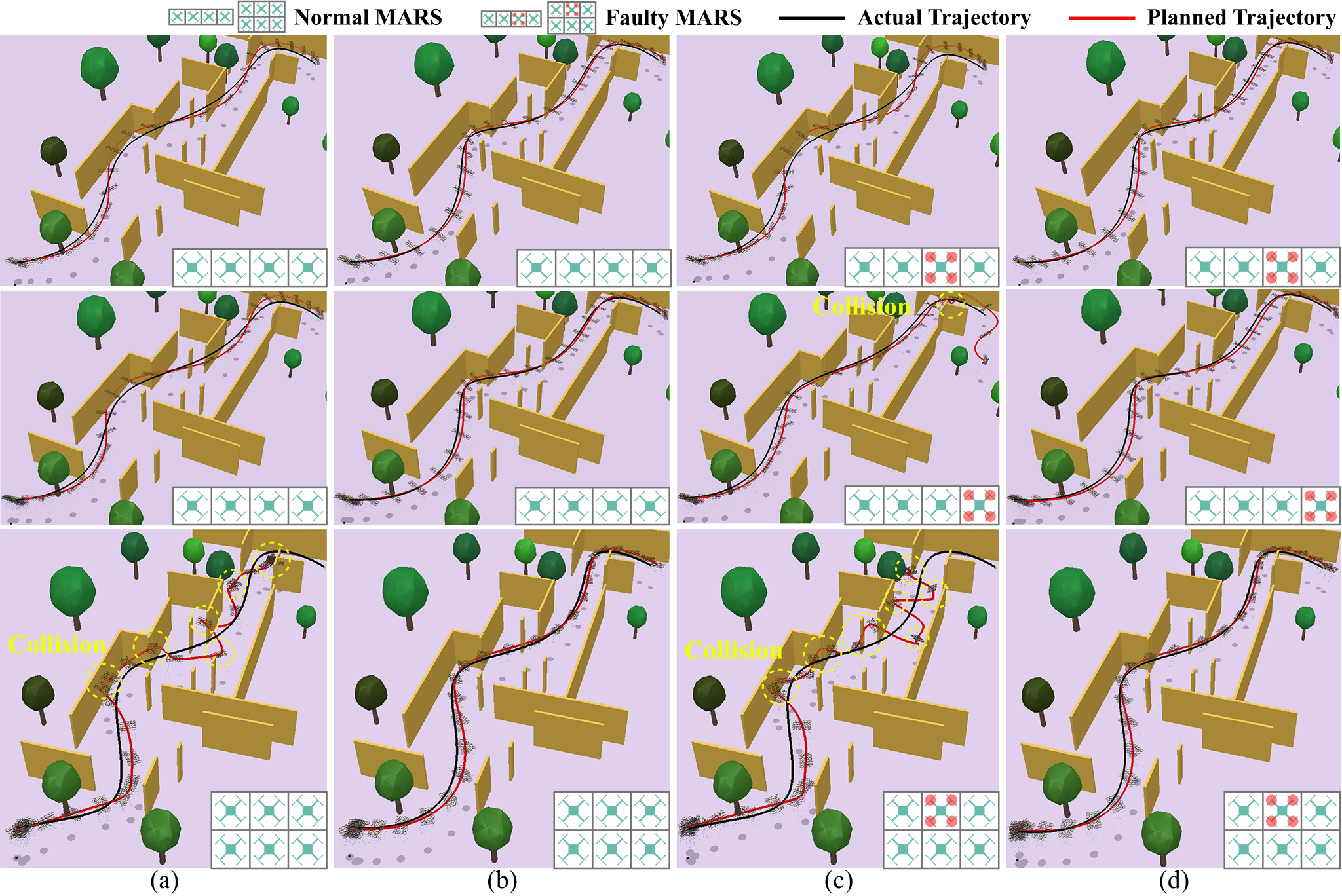}
\vspace{-3mm}
\caption{Simulation of collision-free trajectory planning for $4\times1$ and $3\times2$ assemblies. The annotated diagram illustrates the configuration of MARS, where faulty units are marked in red. The simulation videos are available at \url{https://github.com/RuiHuangNUS/MARS-FTCP/}.  
(a) Normal assembly (SVSDF)~\cite{wang2024implicit}.  
(b) Normal assembly (Ours).  
(c) Post-failure assembly with FTC (SVSDF).  
(d) Post-failure assembly with FTC (Ours). }
\label{fig:trajectory planning result}
\vspace{-3mm}
\end{figure*}
Additional experiments were conducted to evaluate the performance under the simultaneous complete failure of two units, as shown in Fig.~\ref{fig:Complete Failure of One and Two Units} (c)-(e). The tracking errors presented in Fig.~\ref{fig:Complete Failure of One and Two Units} (d1) and (e1) demonstrate greater stability. This improvement is attributed to the symmetric positioning of the faulty units relative to the assembly's center of mass, which results in a more balanced center of gravity. Consequently, this configuration minimizes adverse effects on the roll and pitch control of the assembly. Furthermore, reduced errors are observed in Fig.~\ref{fig:Complete Failure of One and Two Units} (d2) and (e2). The minimal variance further highlights the robustness of our proposed FTC method.

\subsubsection{Rotor Failure and Performance Degradation}
Performance degradation is a common type of malfunction, often caused by propeller or rotor failures. To evaluate the fault-tolerant capability of MARS, we performed trajectory tracking experiments under different rotor failure conditions in a 3$\times$2 configuration, as summarized in Table~\ref{table:Thruster failure 3x2}. Here, $\eta_{2,i}$ represents the coefficient factor of the $i^{th}$ rotor in unit No.~2. The results demonstrate that fault-tolerant control significantly reduces trajectory errors, achieving performance comparable to the fault-free configuration, thereby enabling successful trajectory tracking. Furthermore, when multiple rotors experience reduced thrust, hybrid FTC effectively enhances flight stability. In contrast, the group without internal FTC exhibited considerable trajectory deviations.
\begin{table}
\centering
\caption{Trajectory Error Analysis: Rotor Failures in 3$\times$2 Assembly.\label{table:Thruster failure 3x2}}
\begin{tabular}{c|c|c|c } 
\Xhline{1pt}
 \textbf{Shape} & \textbf{Hybrid FT} & \textbf{Rotor Failure}& \textbf{Error} ($m$)\\ 
\Xhline{1pt}
 3x2& \text{NA}  &\begin{tabular}[c]{@{}c@{}}\text{No failure}\end{tabular}& 0.2230\\
 \hline
 3x2& \ding{56}  &\begin{tabular}[c]{@{}c@{}}$\eta_{2,1}=1, \eta_{2,2}=0.5,$\\ $\eta_{2,3}= 1,\eta_{2,4}=1$\end{tabular}& 0.9050\\
 \hline
  3x2& \checkmark  &\begin{tabular}[c]{@{}c@{}}$\eta_{2,1}=1, \eta_{2,2}=0.5,$\\ $\eta_{2,3}= 1,\eta_{2,4}=1$\end{tabular}& 0.2933\\
 \hline
  3x2& \ding{56} &\begin{tabular}[c]{@{}c@{}}$\eta_{2,1}=0.5, \eta_{2,2}=0.5,$ \\ $\eta_{2,3}= 1,\eta_{2,4}=1 $\end{tabular}& 4.3790 \\
 \hline
   3x2& \checkmark &\begin{tabular}[c]{@{}c@{}}$\eta_{2,1}=0.5, \eta_{2,2}=0.5,$ \\ $\eta_{2,3}= 1,\eta_{2,4}=1 $\end{tabular}& 0.3066 \\
\Xhline{1pt}
\end{tabular}
\vspace{-5mm}
\end{table}
\subsection{Self-Reconfiguration Fault Tolerance Control}

FTC is a crucial component in enhancing the robustness of self-reconfiguration tasks \cite{Huang2025Robust}. In this section, we compare our proposed FTC method with that of \cite{gandhi2020self}. Fig.~\ref{fig:Oscillation fig} illustrates the FTC performance during self-reconfiguration with a faulty unit. We conducted experiments using two different configurations: a $3\times1$ assembly and a $3\times3$ plus assembly.  Fig.~\ref{fig:Oscillation fig} (b) presents the FTC performance during self-reconfiguration in the $3\times1$ assembly. When Unit No.1 separates from the assembly (Fig.~\ref{fig:Oscillation fig} (b1)) and docks at Position No.3 (Fig.~\ref{fig:Oscillation fig} (b2)), only minor oscillations can be observed in the magnified red propeller. In contrast, Fig.~\ref{fig:Oscillation fig} (a1) and (a2) show significant oscillations using the baseline method \cite{gandhi2020self}, indicating that it produces considerable trajectory oscillations during self-reconfiguration (both separation and docking). Such oscillations may lead to potential collisions or safety risks. Additionally, similar oscillations can be observed in the $3\times3$ plus assembly, as shown in Fig.~\ref{fig:Oscillation fig} (d1)–(d6). In contrast, our method significantly reduces oscillations in both the $3\times1$ and $3\times3$ plus assemblies. Moreover, the acceleration and yaw angle profiles recorded during docking and separation show a maximum variation of only 1 degree.  \textbf{This highlights the effectiveness of our proposed adaptive configuration FTC method in enhancing robustness during self-reconfiguration across different unit numbers and configurations.}

\subsection{Collision-free Path Planning}
\begin{figure}[!t]
\centering
\includegraphics[width=3.4 in]{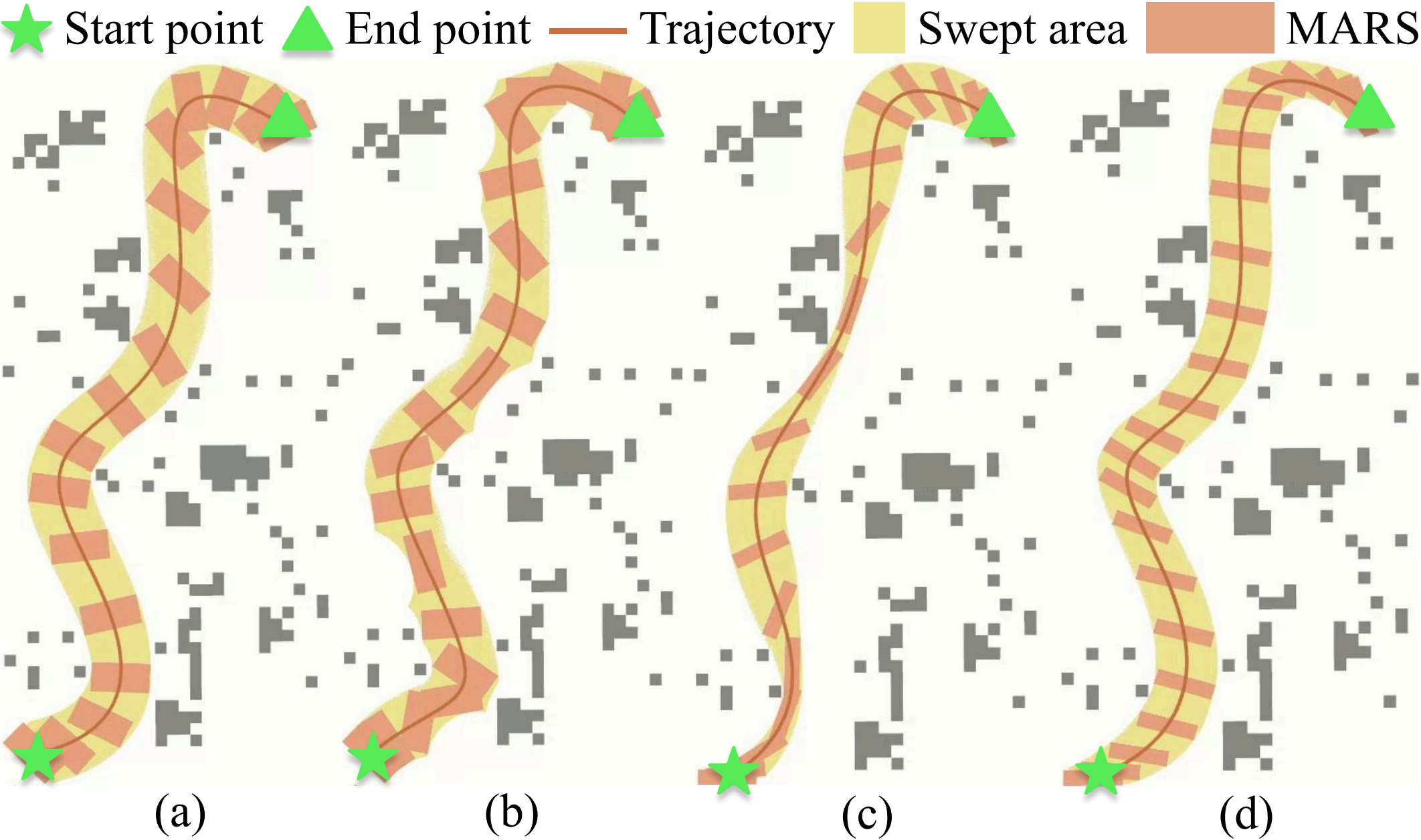}
\vspace{-7mm}
\caption{Comparison of dynamical trajectory generated by SVSDF \cite{wang2024implicit} and our method. (a) 3$\times$2 assembly  (Ours). (b) 3$\times$2 assembly (SVSDF). (c) 4$\times$1 assembly (SVSDF). (d) 4$\times$1 assembly (Ours). }
\label{snake}
\vspace{-3mm}
\end{figure}
To better demonstrate the advantages of the optimal attitude in simulation environments with dynamic models, we carry out experiments using SVSDF\cite{wang2024implicit} and our method in Coppeliasim\cite{coppeliaSim}. The generated MARS trajectories of the 4$\times$1 and 3$\times$2 assemblies are shown in Fig ~\ref{snake}. The attitude-optimized dynamical trajectories minimize abrupt attitude changes and enhance the feasibility of tracking tasks. 

The simulation sequence graphs of different methods are shown in Fig.~\ref{fig:trajectory planning result}. FTC enables MARS to track the planned trajectory after the failure of a unit in both methods, as illustrated in Fig.~\ref{fig:trajectory planning result} (c) and (d). However, the trajectory generated by SVSDF poses greater challenges for MARS with dynamic models, often resulting in more pronounced trajectory errors and a higher frequency of collisions, as shown in Fig.~\ref{fig:trajectory planning result} (a) and (c). Specifically, in the first row of Fig.~\ref{fig:trajectory planning result}, a failure occurs in Unit No.~3 of the $4\times1$ assembly. Our method demonstrates lower tracking errors compared to~\cite{wang2024implicit}. When a failure occurs in Unit No.~4, the increased fault torque reduces the assembly’s control capability, leading to higher tracking errors and even collisions, as shown in the second row of Fig.~\ref{fig:trajectory planning result} (c) for the comparison method~\cite{wang2024implicit}.

Furthermore, for the $3\times2$ assembly, MARS without fault struggles to accurately follow the SVSDF-generated trajectory, as seen in the third row of  Fig.~\ref{fig:trajectory planning result} (a). The performance further deteriorates in the presence of faults, as illustrated in Fig.~\ref{fig:trajectory planning result} (c) in the third row. In contrast, our approach, shown in Fig.~\ref{fig:trajectory planning result} (b) and (d) in the third row, effectively mitigates collision risks and reduces average trajectory errors, achieving performance comparable to the nominal configuration. Table~\ref{table:Collision-free Path Planning} summarizes the trajectory tracking errors shown in Fig.~\ref{fig:trajectory planning result}. Across all configurations, our method achieves improvements of at least 48.5\% and up to 79.5\%, with an average improvement of 69.7\%. \textbf{These results demonstrate the superiority of the proposed method in ensuring dynamic adaptability and collision-free trajectory planning.}
\begin{table}
\centering
\caption{Trajectory Error Analysis: Collision-free Path Planning \label{table:Collision-free Path Planning}}
\begin{tabular}{c||c|c|c } 
\Xhline{1pt}
 \textbf{Shape (Fault ID)} 
 & \textbf{ Baseline\cite{wang2024implicit}} ($m$) 
 & \textbf{Ours} ($m$) 
 &  \textbf{Improvement}\\ 
\Xhline{1pt}
$4\times1$ (Normal)&0.5536 & 0.2849 & $48.5\%$\\
 \hline
$4\times1$ (No. 3) &0.7499 & 0.3039 & $59.5\%$\\
 \hline
$4\times1$ (No. 4) & 0.9923 & 0.3068 & $69.1\%$\\
 \hline
$3\times2$ (Normal) & 1.1777 & 0.3216 & $72.7\%$\\
 \hline
$3\times2$ (No. 2) &1.7007 & 0.3486 & $79.5\%$\\
 \hline
\textbf{Average} & 1.0348 & 0.3132 & $69.7\%$\\
\Xhline{1pt}
\end{tabular}
\vspace{-5mm}
\end{table}
\section{Conclusion}
To achieve robust and safe post-failure control of MARS, this paper proposes a partial and full fault-tolerant power allocation method that redistributes the expected force and torque across all units based on moment arm magnitudes. Compared to existing approaches, our method significantly reduces oscillations during the reconfiguration process. Furthermore, we develop a dynamics-aware, collision-free trajectory planning strategy by extending the A* algorithm to compute optimal attitude nodes that maximize control performance. We then generate a dynamic trajectory incorporating a dynamic feasibility term. Experimental results demonstrate that our approach effectively reduces tracking errors while preventing collisions when following a collision-free trajectory, even in the presence of failures, achieving an average improvement of $69.7\%$.

\bibliographystyle{IEEEtran}
\bibliography{reference}

\end{document}